\newif\ifabstract
\abstracttrue
\abstractfalse 
\newif\iffull
\ifabstract \fullfalse \else \fulltrue \fi

\documentclass[11pt]{article}
\usepackage{amsfonts}
\usepackage{amssymb}
\usepackage{amstext}
\usepackage[ruled,vlined]{algorithm2e}
\usepackage{amsmath}
\usepackage{xspace}
\usepackage{theorem}
\usepackage{graphicx}
\usepackage{url}
\usepackage{graphics}
\usepackage{colordvi}
\usepackage{colordvi}
\usepackage{subfigure}
\usepackage{lscape}
\usepackage{natbib}
\usepackage[english]{babel}
\usepackage{booktabs}
\usepackage{tabularx}

\advance \oddsidemargin by -0.8in
\newcommand{\myparskip}{3pt}
\parskip \myparskip

{\hspace*{\fill}$\Box$\par\vspace{4mm}}

\begin{document}

\title{Modelling Latent Travel Behaviour Characteristics\\ with Generative Machine Learning \footnote{\textbf{Published in the proceedings of IEEE Intelligent Transportation Systems Conference 2018}}}
\author{Melvin Wong\thanks{Laboratory of Innovations in Transportation (LiTrans), Ryerson University, Canada, Email: {melvin.wong@ryerson.ca}} \and Bilal Farooq\thanks{Laboratory of Innovations in Transportation (LiTrans), Ryerson, Canada, Email: {bilal.farooq@ryerson.ca}}}

\begin{titlepage}
	\maketitle
	
	\thispagestyle{empty}
	
	\begin{abstract}
		In this paper, we implement an information-theoretic approach to travel behaviour analysis by introducing a generative modelling framework to identify informative latent characteristics in travel decision making.
		It involves developing a joint tri-partite Bayesian graphical network model using a Restricted Boltzmann Machine (RBM) generative modelling framework.
		We apply this framework on a mode choice survey data to identify abstract latent variables and compare the performance with a traditional latent variable model with specific latent preferences --  safety, comfort, and environmental.
		Data collected from a joint stated and revealed preference mode choice survey in Quebec, Canada were used to calibrate the RBM model.
		Results show that a significant impact on model likelihood statistics and suggests that machine learning tools are highly suitable for modelling complex networks of conditional independent behaviour interactions.
		
	\end{abstract}
	
\end{titlepage}

\section{Introduction}
The increased use of psychological and perceptual variables in travel choice survey have motivated a number of studies that investigated the explicit effects of latent behaviour in decision-making. Analysis of travel mode choice has focused on the effects of modal travel cost, time or reliability and many recent studies have attributed latent behaviour variables to account for unobservable effects  \cite{paulssen2014values,bhat2015introducing}. The Integrated Choice and Latent Variable (ICLV) model is a recent development in structural equation modelling (SEM) to handle hybrid endogenous and exogenous variables in decision-making \cite{ben2002integration}. The ICLV model has been shown -- in some situations -- to produce consistent estimates of model parameters, leading to better explanatory solutions \cite{vij2016and}.
The history of structural modelling dates back to the 1970s and have been originally used in psychology, sociology and market research, and recently  it has seen growing applications in travel behaviour involving latent preference ``attitudinal'' variables and measurement ``indicators''. The fundamental methodology of SEM assumes prior statistical relevance and prior hypothesis about the subjective variables. Errors in measurement and model structure can be independently estimated and psychological effects can be directed using measurement indicators. One of the characteristics of latent variable models is that the estimated model parameters are not always unique. The information quality of the underlying data also poses a significant identification problem in SEM. Theoretical analysis of how latent variables can be identified practically is an important consideration, specifically in the domain of travel behaviour analysis.

Recent studies into some of the insights of decision making process with latent variables have investigated the use of machine learning algorithms to enrich
limited endogenous variables by learning a generative statistical model designed specifically to avoid the problems with non-unique parameter estimates.
Efficient generative modelling algorithms, e.g., RBMs or Variational Autoencoders, developed for machine learning applications, can be applied to latent travel behaviour models without the need for measurement indicators nor through SEM by incorporating choice posteriors to learn latent variable interactions \cite{wong2017discriminative}.

The integration of machine learning algorithms in econometric models have constituted a substantial research topic in recent years to uncover underlying anomalies in random utility maximization (RUM) theory \cite{rosenfeld2012combining}.
Recent advances in generative modelling techniques have fuelled interest in analysis in latent variable models and distributed representations of latent variables in undirected models \cite{hinton2006fast}.
They been used to provide data density estimation, inference, information retrieval and multi-class classification in both supervised and non-supervised setting \cite{larochelle2012learning}.
It is also possible for machine learning algorithms to be used for extracting information priors, as long as the learning parameters are bounded to specific behaviour constraints \cite{rosenfeld2009modeling}.
The proposed method is similar to variational inference, by using a layer of hidden units to learn non-linear latent representation of the observed data.

In this paper, we aim to develop a novel conditional RBM (C-RBM) for travel survey data that can leverage attitudinal and causal information of choice preference simultaneously.
The main contributions of this paper are:
\begin{itemize}
	\item Propose a C-RBM framework for travel behaviour model to incorporate conditional relationship between observed and latent information.
	\item Explore the capability and identifiability of the framework for latent behaviour characteristics.
	\item A empirical comparison with traditional SEM based discrete choice model.
\end{itemize}

\section{Literature Review}
\subsection{Structured Equation Modelling (SEM)}
The use of attitudes and perceptions in latent variable modelling have been used in various implementations and approaches in travel behaviour models \cite{daly2012using,ashok2002extending}.
The ICLV model is a particularly useful SEM method which incorporates psychometric indicators by constructing a model in terms of a system of unidirectional effects of one variable to another \cite{golob2003structural}.
Within this domain, ICLV models estimate either sequentially or simultaneously on latent variables and indicator manifestation to explain utility of each alternative.
The ICLV model combines consideration for unknown variables with the choice model, offering better explanatory effect.

Early developments of latent behavioural framework are a response to the need for interactions between psychometric data and choice preferences, treating behaviour as an ``open black-box'' \cite{morikawa2002discrete}.
A distinction in SEM is in the effects between observed and target perceptual variables are pre-specified graphically.
The direct effects of indicator measurements on latent variables are expected to be available.
For instance, survey collection does not take into account psychometric factors, latent variables cannot be estimated using the ICLV method.
Even when measurement indicators are available they may be weak predictors of latent variables if the respondents do not understand or inaccurately answer those questions.
Indicators may also not provide further useful information and might cause mis-specification of the choice model \cite{vij2016and}.
ICLV explicitly models unobserved (latent) behaviour factors through measurement equations.

The primary assumption of discrete choice models is that the unobserved choice processes are implicitly captured by the model \cite{ben2002integration}.
Representing decision makers underlying psychological and sociological reasons as so called latent variables is that while observed characteristics may explain certain direct choice behaviour, the confounding effects still remain subjective.
\subsection{General specification of the ICLV model}
The ICLV choice model is composed of 3 sub-parts: The choice model, the measurement model and the latent variable model \cite{vij2016and}.
A maximum likelihood estimation (MLE) function is used to estimate the parameter values.
The observable variables consists of generic and alternative specific inputs where all the respondents $n$ gives their stated choice preference $i$. The set of inputs are referred to as $x_m$ and $x_{im'}$ respectively.
In a standard RUM-based multinomial logit utility, the utility is defined by:
\begin{equation}
U_i = V_i + \varepsilon_i = \beta_m x_{im'} + \beta_{im} x_m + \varepsilon_i,
\end{equation}

\noindent where $\beta_{m}$ and $\beta_{im}$ are the parameters that define the sensitivity of each variable and $\varepsilon_i$ is the extreme valued error term.
$V_i$ represents the observed part of the utility.
For simplicity, we assume that the alternatives are homogeneous across the population and parameters are estimated without a variance parameter.

The latent variable model extends the utility by adding a latent variable term $x_h^*$ where (h) is the number of latent variables required.
An equation is defined for each latent variable.
Random utility with latent variable can be defined as follows:
\begin{equation}
U_i = V_i + \beta_{ih} x_h^* + \varepsilon_i.
\end{equation}

Typically, the functions for latent variable are not explicitly defined beforehand.
Here we provide several possible ways of how the latent variable can be formulated in terms of observable variables.
The measurement model decouples the latent underlying factors from the observed variables and separate representations into discrete, measurable points, e.g. latent attributes such as `attitude towards owning a car'.
Indicators $I$ define the response of the individual to perceptual questions.
For instance, one can ask the question `What is the importance of safety when choosing to travel by train?'
The response is usually defined by a Likert scale, we assume that all indicators are configured as binary valued $\left[0, 1\right]$ (e.g. not important or important).
Similarly in latent variables, each indicator is defined with one equation.
Each equation measures the distribution of indicators conditional on the values of the latent variables,  $f(x_h^*)$.
For example, indicators can be defined as a conditional probability distribution of latent variables:
\begin{equation}
I_j = \beta_{jh} x_h^* + \varsigma_j,
\end{equation}

\noindent where $\varsigma_j$ represents the error terms of the indicators and $\beta_{jh}$ is the parameters defining the weight of the latent variable on the specified indicator.
The parameters $\beta_{jh}$ can be estimated by the probability that the indicator is $I=1$, using a binary logit model:
\begin{equation}
p(I_j=1|x_h^*) = \frac{e^{\beta_{jh} x_h^*}}{\sum_{j' \in \left[0, 1\right]} e^{\beta_{j'h} x_h^*}}
\end{equation}

In principle, any function for $I$ is possible (including linear when the indicator is a scale, hence a Probit model), but we consider a logit function for simple generalization.
Providing indicators may help to capture the systematic response bias not found in observed variables.
However, this method cannot be used if psychometric indicators are not available.

\subsection{Modelling Non-linearity in Latent Variables}
Non-linear interaction terms between latent and observed variables allows for cases where latent variables are not monotonically related to observed variables.
The difficulty in computing the covariance or correlation matrices among non-linear terms of exogenous latent variables limits the use of non-linear functions, thus requiring non-linear constraints which results in an increase in complexity of the model specification and identification.
The linear function used in \cite{vij2016and} is described in ICLV models \cite{bhat2015introducing}:
\begin{equation}
x_h^* = f(x_m) = \beta_{hm} x_m + \vartheta_h,
\end{equation}

\noindent where $\beta_{hm}$ is the parameter describing the linear relation between observed and latent variables and $\vartheta_h$ is a random stochastic term.
This form is selected because the function will be linear and continuous and can be easily inferred from.
However, there is a risk of overestimation as the value of $x_h^*$ is not bounded $(x \in \left[ -inf, inf \right])$.
There would be potential numerical instability in the gradient estimation procedure (when taking the exponential of a large number input).
A common practice in discrete choice modelling to stabilize parameter identification is to scale the input values to a small (<1.0) number or include a scale estimator \cite{klette1996inconsistency}.

A non-linear formulation is the sigmoid or inverse logit function $f(x) = sigmoid(x) = \frac{1}{1+ e^{-x}}$.
This is common for latent variables in discrete choice models since the output is continuous and bounded between 0 and 1.
Intuitively, the value of the latent variable will represent a probability that the latent variable is available in the choice. The formula is as follows:
\begin{equation}
x^*_h = sigmoid(\beta_{hm} x_m) = \frac{1} {1 + e^{-\beta_{hm} x_m}}
\end{equation}

Other possible functions of $f(x_m)$ include the rectifier model (commonly referred to as $Relu(x)$ in machine learning literature) is a threshold version of the linear function with $f(x) = Relu(x) = max(0, x)$ and the soft rectifier $f(x) = softplus(x) = \ln(1+e^x)$ where $x = \beta_{hm} x_m + \vartheta_h$ (Glorot et al. 2011).
When $x = x_{m_1} - x_{m_2}$, the resulting output becomes a measure of alternative regret.


\section{Framework and Estimation of latent variable model through C-RBM algorithm} \label{framework}
Generative models learn the underlying choice distribution $p(y)$ and latent variable distribution $p(x^*|y)$ given some input variables $x$.
A Bayesian inference method is used to derive the posterior distribution of $y$ given some observed and/or latent variable, e.g. $p(y|x^*) = \frac{p(x^*|y)p(y)}{p(x^*)}$.
Latent variables are features which perform non-linear generalization of the highly heterogeneous observed data.
Intuitively, in terms of econometric analysis, latent variables in generative models are arbitrary variables that depend on observed data, including response choices.
In ICLV models measurement functions may be prone to errors.
This is not so in the case of generative models, as latent information is inferred from choice data (through a Markov network for example).
The C-RBM is a variant of a Boltzmann machine inference model with an undirected energy-based model (from the basis of information theory and relative entropy) and a tri-partite of variables having symmetric connections.

The RBM framework estimates the the amount of information `bits' required to map the data onto the set of latent variables.
In addition, each group is conditioned on another set of inputs, in the case of an ICLV, the observed variables can be used as conditional inputs.
The latent variables are assumed to be independent of each other and the model has stochastic visible variables $y \in \lbrace 0, 1\rbrace \forall \mathcal{Y}$ and latent variables $h \in \lbrace 0, 1\rbrace^J$ conditioned on some known prior distribution $x$.
In discrete choice modelling, one of each constants or alternative specific parameters is fixed to zero.
This can be performed in stochastic gradient learning by setting the gradient update to zero of the associated parameter in the computational graph.

The joint distribution of visible and latent variables is given by the Hopfield energy function:
\begin{align}
	Energy(y, x^*, x) &= \sum_{i \in I} y_i c_i - \sum_{j \in J} x^*_j c_j - \sum_{i,j} x^*_j D_{ij} y_i \nonumber\\
	&- \sum_{i \in I} x_{im} B_{i}  - \sum_{j \in J} x_{m} G_{hm}
\end{align}
\noindent where $c_i$ and $c_j$ are the constant values associated with the alternatives and latent variables respectively.
$D_{ij}$ is the parameter covariance matrix representing the relation between the latent and alternatives.
$B_i$ is the parameter vector of the conditional alternative specific inputs $x_{im}$.
$G_{hm}$ is the parameter matrix expressing the relation between latent and observed generic variables, likewise one parameter vector row is fixed to zero for model identifiability.
We can express the Boltzmann distribution as an energy model with energy function which relates the entropy of the model to a specific state of the machine$F(y)$:
\begin{equation}
p(y) = \frac{1}{Z} \sum_{x^*} exp(-F(y))
\end{equation}

where $Z$ is the partition function $Z = \sum_{i,j} \exp(-Energy(y,x^*,x))$ over all possible latent vector combinations. $F(y)$ is defined as the \textit{free-energy} function:
\begin{align}
	F(y) &= - \ln \sum_{x^*} \exp(-Energy(y,x^*, x)) \\
	F(y) &= -y_i c_i - \sum_{j \in J} \ln(1 + \exp(D_{.,j} y + c_j))
\end{align}

\subsection{Objective function and likelihood estimation}
To estimate a ICLV model, maximum likelihood (ML) is used most often.
ML maximizes the probability that the structural model parameters generates the implied output and the measurement model maximizes the probability that the underlying latent variables generates the associated indicators.
To perform estimation of RBM type models, we need to define the objective that is robust and stable in the biases of the standard errors.
A stochastic graph is constructed that incorporates both conditional dependence and the choice model.
The C-RBM model learns aspects of an unknown probability distribution based on samples from that distribution.
A stochastic gradient descent algorithm iterates across all observations and updates the parameter vectors such that the model best represent the distribution of the choice data (Algorithm \ref{algocrbm}).
To generate latent variables, it is necessary to compute the log likelihood of the joint distribution $p(y, x^*, x)$.
Efficient Markov Chain Monte Carlo algorithm have been developed to deal with such problems using Gibbs chain sampling methods and contrastive divergence (CD).
Assuming that individual responses are known, we can model the joint distribution of the responses and latent variables using the Bayesian estimation rule:
\begin{equation}
p(y) = \int_{x^*} p(y, x^*, x) dx^*  = \int_{x^*} p(y|x^*, x)p(x^*|x) dx^*
\end{equation}

The probability that the C-RBM model estimates is based on comparing the Kullback-Leibler divergence of the initial probability distribution $p(y)$ and another, final distribution $p(\hat y)$, where $p(\hat y)$ is the probability of the reconstructed representation after Gibbs sampling.
To find the gradient derivative for the gradient descent training algorithm, we take the derivative of the log probability of the training vector with respect to the model parameters:
\begin{align}
	\frac{\delta \log p(y)}{\delta \theta} &= <y_i x^*_j>_{data} - <y_i x^*_j>_{model}\nonumber\\
	&= \phi^+ - \phi^-
\end{align}

\noindent where the components of $\langle y_i x^*_j \rangle$ corresponds to the expected value under the specified distribution (data or model).
The first and second terms are the positive and negative phases of the Gibbs sampling procedure.
The update rule from the model parameters can be performed with stochastic gradient descent (SGD) at each iteration $t$:
\begin{align}
	\Delta \theta &= \Phi (<y_i x^*_j>_{data} - <y_i x^*_j>_{model})\\
	\theta_t &= \theta_{t-1} - \Delta \theta
\end{align}

\noindent We incorporate a learning factor $\Phi$ in the objective function which controls the magnitude of the update parameters.
The objective assumes that the marginal $p(x^*|x)$ has a closed form solution and the function generate output samples $\hat y \sim p(\hat y=1|x^*, x)$.

\begin{algorithm}[!t]
	\caption{Conditional RBM Gibbs sampling procedure using Contrastive Divergence}
	\label{algocrbm}
	\SetAlgoLined
	\SetKwInOut{Input}{Input}
	\SetKwInOut{Output}{Output}
	
	\Input{
		Data sample $\mathcal{D}$,
		batch sample $S_i\subset\mathcal{D}$, $i=1,...,s$,
		iteration steps $T$}
	\Output{Model parameters $\theta$.}
	\BlankLine
	initialize: $\theta=0$\;
	\ForAll{$S_i\in\mathcal{D}, \tau=1,...,T$}
	{
		\ForAll{$(y,x^*, x)\in S_i$}
		{
			\For{$n=1$ \KwTo $N$}
			{
				iterate over Gibbs chain, CD$_n$ \\
				$<y,x^*>_{data} \leftarrow p(y_n, x^*_n, x_n)$\\
				\BlankLine
				Sample: $\hat y \sim p(y|x^*,x)$ \\
				$<y,x^*>_{model} \leftarrow p(\hat y_n, x^*_n, x_n)$\\
			}
		}
		parameter update:\\
		{$\Delta \theta \leftarrow \Phi(<y, x^*>_{data} - <y, x^*>_{model})$ \\}
		\ForAll{$\theta$}
		{
			$\theta_{\tau} \leftarrow \theta_{\tau - 1}
			-\Delta \theta$\;
		}
	}
\end{algorithm}

\subsection{Construction of the latent behaviour choice model}
The generated parameter vectors of the C-RBM model are then used to estimate a latent behaviour model that contains the utility maximizing estimator for each observed and latent variables with an indicator model for the latent variable component:
\begin{itemize}
	\item For the choice model y, the estimator simply calculates the likelihood $\mathcal{L}(\theta)$ under the RUM theory, that is $\mathcal{L}(\theta) = \frac{1}{n} \sum_n p(y|x^*, x; \theta)$. Here $\theta$ are the parameters of the generic ($\beta_{im}$), alternative specific ($\beta_i$) and latent ($\beta_{hi}$) variables. \\
	\item For the latent variable $x^*$, we calculate the conditional probability $p(x^*|x)$.
	A reparameterization boundary condition is placed on latent variables $\left[0, 1\right]$. The identified parameter represents the probability that the latent variable is present in the individual. \\
	\item For the indicator component, statistically significant latent variables are extracted from our C-RBM model estimation. \\
\end{itemize}

The choice model be can of any form, e.g. multinomial logit, mixed logit, nested logit, or a combination of different choice mechanisms (for simplicity, we use a MNL in our experiment).
Once choice and measurement model are formulated, the likelihood function is derived to estimate the parameters of the model.
The likelihood function is defined as the mixed logit integral of the choice model conditional on the indicator measurement model:
\begin{equation}
P(y|x,x_h^*,I) = \int P(y|x,x^*)P(I|x^*) dx^*
\end{equation}

Assuming that the measurement model follows the logistic sigmoid function with scale and/or translation factor, the integral can be estimated by maximum log likelihood (MLE) and terms of the resulting densities are:
\begin{align}
	L(\theta) &= \log(P(y|x,x_h^*,I)) \nonumber\\
	&= \sum_n ( \log(P(y|x,x^*)) + \sum_j \log(P(I|x^*)) )
\end{align}

\noindent The first term is the log likelihood of the choice model. The second term, can be substituted with cross-entropy (CE) maximization:
\begin{equation}
\log(P(I|x^*)) = I * \log(f(x^*|x)) + (1- I) * \log(1 - f(x^*|x))
\end{equation}

The CE approach for multinomial logit models is equivalent to the standard log likelihood for standard discrete-continuous choice models when more than 1 alternative are selected.
In the case of the latent variable and indicator function, the probability of $I_j = 1$ is independent of other $I_{j'}$.
This CE expectation maximization procedure on a multi-attribute logistic function recovers the likelihood of the indicator model efficiently and directly evaluate $P(I|x^*)$ simultaneously with the choice model.

\section{Case study}
\begin{figure}
	\centering
	\includegraphics[width=0.5\textwidth]{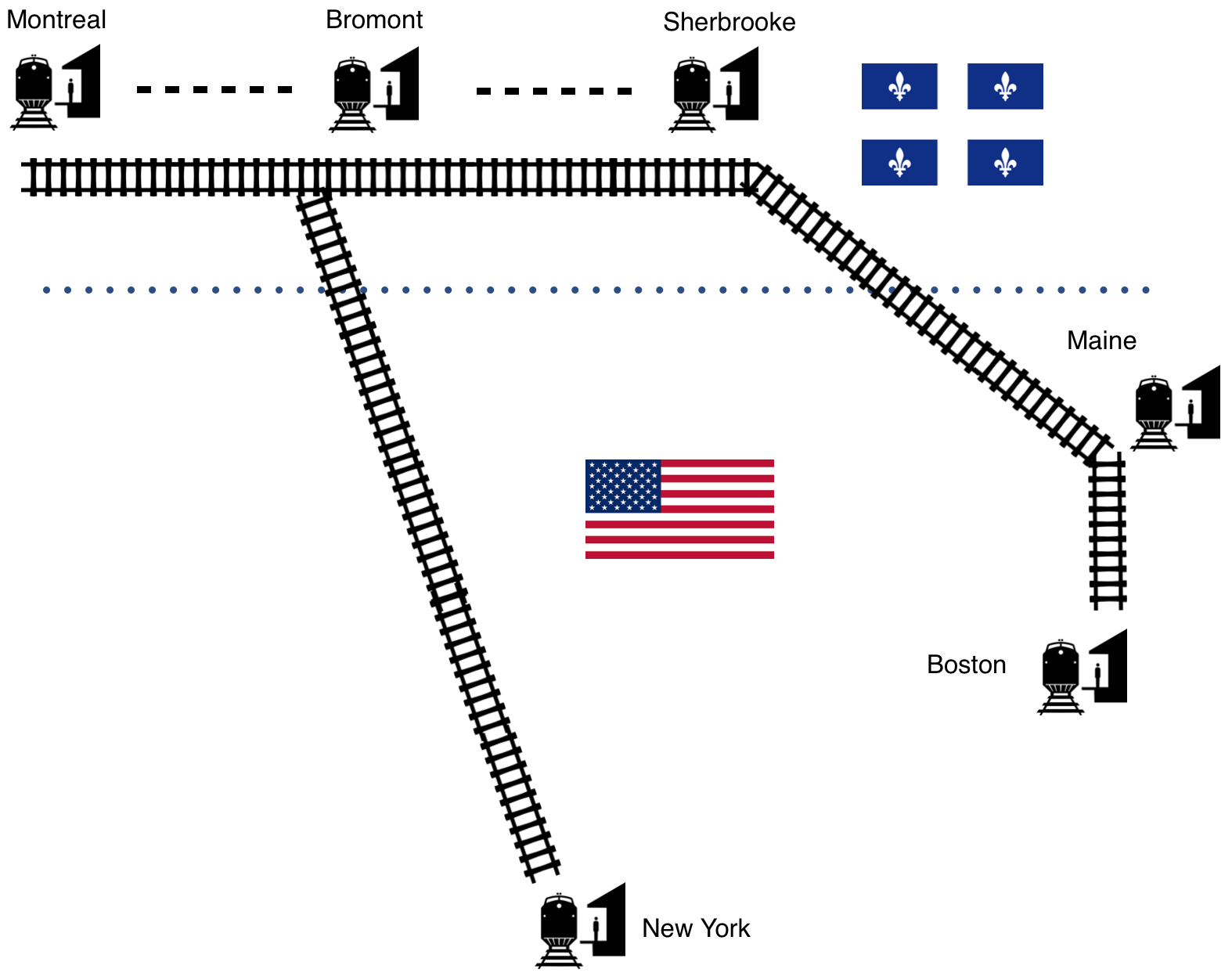}
	\caption{Origin Destination nodes modelled for computational experiments.}
	\label{map}
\end{figure}

A combined revealed and stated preference travel survey from commuters along the Northeastern USA rail corridor with {Montreal} in Canada (\textit{Montreal}, \textit{NYC}, \textit{Maine}, \textit{Boston}) is conducted.
A sleeper train between these cities and tourist destinations (\textit{Train Hotel}) was proposed to provide an alternative to the regular rail travel mode.
The proposed \textit{Train Hotel} provides overnight sleeper amenities and entertainment for round-trip journeys shown in Fig. \ref{map}.
A joint RP-SP survey design provides multi-attributed and generic variables, resulting in more accurate outcomes.
The survey analyses mode choice preference of passengers who travelled between select Canada and USA destinations within 12 months prior from the day of survey.
The data statistics and collection procedure are described in \cite{sobhani2017innovative}.
People who have not travelled to any of the destinations within the time frame were included in the survey to provide a representative distribution of the population in the region that do not often use the rail corridor for commute.
The demand for the new intercity travel mode was estimated for both people who travel between the destinations and those who did not travel but are interested in making a future journey along the route.

\begin{figure}
	\centering
	\includegraphics[width=0.3\textwidth]{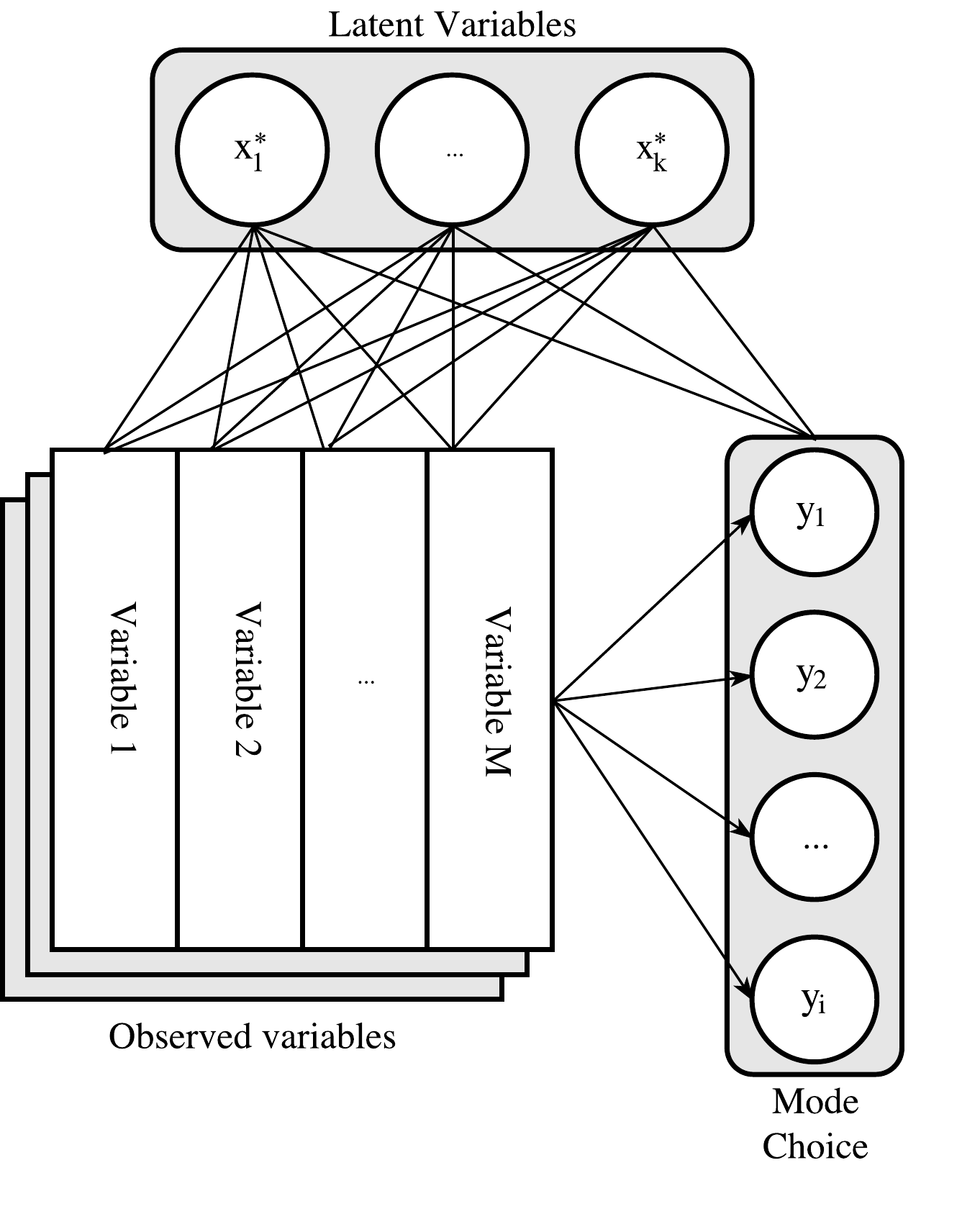}
	\caption{A joint tri-partite RBM structure for travel analysis with latent behaviour variables.}
	\label{configuration}
\end{figure}

In the SP choice survey, each respondent was presented with up to 6 alternatives $y_i \in \lbrace \textit{Bus}, \textit{Car Rental}, \textit{Car}, \textit{Plane}, \textit{Train Hotel}, \textit{Train}\rbrace$ and each mode alternative was characterized by: trip duration, trip reliability and trip cost.
Each attribute was sampled on different levels for each respondent (eg. multiple price levels) defined relative to the origin and destination pairs.
The level of each quantity was randomized across variables to control for potential ordering bias, however the choice order between respondents was not varied.
The second part of the survey data collected socio-economic and household characteristics of the respondents.
The survey data consisted of continuous (e.g. income, age, number of vehicles) and categorical variables (e.g. education, household type).
For consistency, all generic variables related to the respondents' characteristics were binary coded (continuous variables are first categorized).
The model structure used in the analysis is shown in Fig. \ref{configuration}.

For the measurement model, 3 qualitative indicators were considered for each mode: environmental, comfort, safety, (e.g. safety of car, safety of plane).
Respondents indicated their perception of these indicator by level of importance on a 5-point Likert scale.

\subsection{Experiment Settings}
The main characteristics of the study was that commuter travel between Montreal and Northeastern USA destinations generated large amounts of land and air traffic due to the high commercial, leisure and industrial activities along the travel corridor.
When analysing latent behaviour effects, the following factors are considered: the number of latent variables required, inferrability of latent variables, model identifiability, optimization methods and computational speed.

In the C-RBM framework, our model functions are configured as a probabilistic graphical model and all parameters were estimated for the influences on the choice probability.
Alternative specific variables, e.g. cost time and reliability were not part of the latent variable function (since it was not an observed characteristic of the individual).
Generic variables, corresponding to demographic information was used to develop the latent variables.
Alternative specific input variables for cost and travel time were statistically significant in our initial analysis estimated with a MNL model.

Typically, in ICLV models, latent variables were defined prior to model estimation.
This presented a subjective view of personal traits.
In the C-RBM framework, parameters were defined for all variables by minimizing the latent variable to choice reconstruction error.
This offered a significant advantage -- the latent variables would be defined by a known prior distribution.
Following which, the Hessian matrix was computed after each optimization and the statistical effects analysed.
Our results suggest that among the latent variables, several variables have meaningful properties and were semantically sound.
Each latent variable would be classified according to their dependencies (observed generic variables).
This was further influenced by the role of the latent variables explaining the choice patterns.

\section{Results}
The performance for the mode choice were compared as follows: First, we initialize a set of parameter values using C-RBM method.
Next, we constructed a ICLV model with interaction terms (ICLV) using the significant parameters.
Finally, we estimated a latent behaviour model using the C-RBM values as initial starting point (C-RBM).
By initializing from a optimal non-zero point, we can avoid identifiability problems by having a higher probability of finding the global optima through the gradient estimation parameter search.
For estimation using a stochastic gradient method, we fix the gradient for the reference parameter to zero so updates are not backpropagated to the parameters therefore a reference value could be found.

\subsection{Latent Behaviour Model Formulation}
We measure the reliability of the latent variable parameters by quantifying perceptual meaning (e.g. quality measure, attitudes towards a particular habit) to each latent variable that could be used as additional explanatory variable in order to obtain better fit on the choice model.
The latent variables were then evaluated on their consistency through the measurement indicator model.
Through this process, latent variables were hypothesized in a less subjective manner since, they were learned through the C-RBM model framework.
We use significant latent variables as a guide for construction of the ICLV model assuming that there should be a relation between the posterior choice and prior distributions.
Using observed distribution of choice data instead of pre-defined latent variables in our estimated model removes assumed causal relation with subjective indicators.

The equations of the ICLV model follows a 6 alternative mode choice model ($i$) with three latent variables ($x_1^*, x_2^*, x_3^*$), the three latent variables were measured by specific mode indicator variable (e.g. $I_j \forall J \in$ $\lbrace$ \textit{bus}, \textit{car}, \textit{train}, \textit{plane} $\rbrace$).
The measurement indicators $I_j$ were binary coded from a 5-point Likert scale (1, 2, 3 = not important (1), 4, 5 = important (0)) \cite{sobhani2017innovative}.
Latent variable interaction terms, denoted by the observed variables are formulated as:
\begin{itemize}
	\item Environmental \\
	Variables: Driving Licence, Age 25-45, FT workers, HS Education, 0 HH Vehicles, 0 or 2 Children, Income $\geq$60K
	\item Safety \\
	Variables: Public Transit Pass, Age 25-45, 1 HH Vehicle, 0 or 1 Children, 20K$<$ Income $\leq$60K
	\item Comfort \\
	Variables: Age $\geq$45, Male, Tertiary Education, 1 or more HH Vehicles, Income $\leq$20K, Income $\geq$60K
\end{itemize}

Measurement equation of the ICLV model:
\begin{equation}
I_j = f (-\beta_{jh} x_h^*)
\end{equation}

Structural equation of the ICLV model:
\begin{equation}
U_i = \sum_m \beta_i x_{im} + \sum_h \beta_{ih} x_h^* + c_i
\end{equation}

\begin{table}[!t]
	\caption{Alternative Specific Variables and Constants Estimated Within the ICLV and C-RBM Models}
	\centering
	\resizebox{0.5\textwidth}{!}{
		\begin{tabular}{l l l l l l l}
			\toprule
			& \multicolumn{3}{c}{ICLV}                             & \multicolumn{3}{c}{C-RBM}                 \\
			\midrule
			Parameters          & value                    & std. err.                 & t-test  & value     & std. err. & t-test  \\
			\midrule
			ASC\_Bus            & -2.485                   & 0.204                     & -12.179 & 0.266     & 0.209     & 1.273   \\
			ASC\_CarRental      & -2.243                   & 0.33                      & -6.802  & 1.319     & 0.335     & 3.934   \\
			ASC\_Car            & 0.643                    & 0.115                     & 5.619   & 1.236     & 0.118     & 10.507  \\
			ASC\_Plane          & -0.318                   & 0.208                     & -1.531  & -0.779    & 0.209     & -3.732  \\
			ASC\_TrH            & -0.386                   & 0.097                     & -3.967  & -0.452    & 0.098     & -4.609  \\
			ASC\_Train          & 0 (ref.)                 & -                         & -       & 0 (ref.)  & -         & -       \\
			cost                & -0.609                   & 0.112                     & -5.447  & -0.595    & 0.114     & -5.217  \\
			travel time        & -0.127                   & 0.023                     & -5.477  & -0.131    & 0.024     & -5.541  \\
			reliability         & 0.249                    & 0.684                     & 0.364   & 0.42      & 0.692     & 0.606   \\
			\midrule
			\textit{Model statistics}                                                                                              \\
			Null Loglikelihood  && -2917.752                                           &         && -2917.752                       \\
			Final Loglikelihood && -2013.685                                           &         && -1946.872                       \\
			rho square          && 0.310$^*$                                           &         && 0.332$^*$                       \\
			AIC                 && 4273.371                                            &         && 4139.744                        \\
			BIC                 && 4948.5                                              &         && 4814.873                        \\
			\bottomrule
			\multicolumn{7}{l}{\footnotesize $^*$ \textit{note that the functions governing the relationship between} $y$ and $x,x^*$ \textit{are different,}} \\
			\multicolumn{7}{l}{\footnotesize \textit{  so we cannot compare rho square values directly}}                                                         \\
		\end{tabular}
	}
	\label{alt_spec_vars}
\end{table}

\subsection{Model analysis}
The results of the two-stage approach showed that three attitudinal variables can be included for estimating latent behavioural aspects for travel mode between Montreal and Northeastern USA destinations.
The signs of the indicator parameters are as expected in the C-RBM model and the t-tests show that most parameters are significant.
In the ICLV model the high positive values of the latent variable parmeters indicate that individuals are most sensitive to the comfortability of train mode, likewise for environmental conscious behaviour, improving environmental impact of train mode also have the highest impact on perception, while car and plane mode had the least effect.
For C-RBM, experiments on different number of latent variables also showed convergence and identification problems as some latent variables were found to be identical or very similar.
The results for the estimation of SP variables (cost, time and reliability) are shown in Table \ref{alt_spec_vars}.
Both models are consistent in cost, travel time and reliability parameter values.

One can also examine the models behaviourally.
Note that incorrect use of latent variables may lead to inconsistent parameter estimation, but this is generally difficult to avoid as there are many non-linear parameters involved.
Assuming similar parameter estimates, the latent constructs observed in the C-RBM models indicated that effects of choice on latent variables lead to a more accurate representation of behaviour. Comparing the standard ICLV and C-RBM method, there is greater model fit when the parameters are initialized well prior to model estimation.

Our case study reveals the feasibility of the C-RBM framework on mixed RP and SP data which could account for perception effects related to SP values and attitudinal questions.
An important advantage of this is to be able to estimate the values from the data instead of postulating them.
However, we should mention that neither method is fully reliable, but provides a different perspective that is representative of the underlying latent variables.
The statistical results show that our method has superior performance in terms of estimated log-likelihood (-1946.872 vs. -2013.685).
This shows that alternative specific variables do not have any large variance when incorporating latent variables;
Estimating the model through a joint estimation method do not generally influence the underlying factors of alternative dependent cost, time and reliability variables.

Finally, we should note that parameter values cannot be compared directly.
Under assumption that the latent behaviour function is non-linear and complex, there may be multiple local optimal solutions and we did not consider scale and translation effects of the underlying variables and different decision rules in this study.
We are currently investigating these effects in our future work.
Furthermore, these are important considerations when the structure of the latent variable model is changed from a conditional to a joint model.

\section{Conclusion}
In this paper, we develop a new approach to the problem of modelling latent behaviour through estimation a joint distribution from its associated choice and auxiliary information.
This approach has been studied in different contexts in machine learning models recently.
Our C-RBM approach is the first fully developed solution to latent behaviour models.
This approach is comparable with previously developed ICLV methods in terms of model fit and do not require additional parameters.
The estimation process is straightforward and convergence is fast for large parameter vectors using stochastic gradient descent with CD objective function.

In a sample of a new travel mode choice, survey respondents were asked to indicate their preference of travel mode given that a hypothetical intercity train service \textit{Train Hotel} is offered.
\textit{Train Hotel} provides overnight sleeper amenities as an alternative to day trains and other modes such as cars, plane and buses.
Results obtained from the C-RBM parameter estimation are compared with results from the ICLV model.

The ICLV approach analytically derive each latent variable under assumptions on the measurement functions.
This method is effective when indicators are available, but assumptions may be hard to verify as we are unsure about the interactions of the latent variable generating process.
In some cases, theoretical result only gives asymptotic guidance in finite observations.
It is likely that the stated indicators may not be reflective of real attitudes and perception and heavily influenced by the survey conditions, geographic area, socio-demographics or other revealed information.
While these model design choices are data-reliant, having a reliable estimate is a requirement for strong econometric plausibility.
In our case study, inference on model performance is a straightforward task of analysing parameter validity.

\bibliographystyle{plainnat}
\bibliography{bibliography}

\begin{thebibliography}{15}
\providecommand{\natexlab}[1]{#1}
\providecommand{\url}[1]{\texttt{#1}}
\expandafter\ifx\csname urlstyle\endcsname\relax
  \providecommand{\doi}[1]{doi: #1}\else
  \providecommand{\doi}{doi: \begingroup \urlstyle{rm}\Url}\fi

\bibitem[Ashok et~al.(2002)Ashok, Dillon, and Yuan]{ashok2002extending}
Kalidas Ashok, William~R Dillon, and Sophie Yuan.
\newblock Extending discrete choice models to incorporate attitudinal and other
  latent variables.
\newblock \emph{Journal of marketing research}, 39\penalty0 (1):\penalty0
  31--46, 2002.

\bibitem[Ben-Akiva et~al.(2002)Ben-Akiva, Walker, Bernardino, Gopinath,
  Morikawa, and Polydoropoulou]{ben2002integration}
Moshe Ben-Akiva, Joan Walker, Adriana~T Bernardino, Dinesh~A Gopinath, Taka
  Morikawa, and Amalia Polydoropoulou.
\newblock Integration of choice and latent variable models.
\newblock \emph{Perpetual motion: Travel behaviour research opportunities and
  application challenges}, pages 431--470, 2002.

\bibitem[Bhat et~al.(2015)Bhat, Dubey, and Nagel]{bhat2015introducing}
Chandra~R Bhat, Subodh~K Dubey, and Kai Nagel.
\newblock Introducing non-normality of latent psychological constructs in
  choice modeling with an application to bicyclist route choice.
\newblock \emph{Transportation Research Part B: Methodological}, 78:\penalty0
  341--363, 2015.

\bibitem[Daly et~al.(2012)Daly, Hess, Patruni, Potoglou, and
  Rohr]{daly2012using}
Andrew Daly, Stephane Hess, Bhanu Patruni, Dimitris Potoglou, and Charlene
  Rohr.
\newblock Using ordered attitudinal indicators in a latent variable choice
  model: a study of the impact of security on rail travel behaviour.
\newblock \emph{Transportation}, 39\penalty0 (2):\penalty0 267--297, 2012.

\bibitem[Golob(2003)]{golob2003structural}
Thomas~F Golob.
\newblock Structural equation modeling for travel behavior research.
\newblock \emph{Transportation Research Part B: Methodological}, 37\penalty0
  (1):\penalty0 1--25, 2003.

\bibitem[Hinton et~al.(2006)Hinton, Osindero, and Teh]{hinton2006fast}
Geoffrey~E Hinton, Simon Osindero, and Yee-Whye Teh.
\newblock A fast learning algorithm for deep belief nets.
\newblock \emph{Neural computation}, 18\penalty0 (7):\penalty0 1527--1554,
  2006.

\bibitem[Klette and Griliches(1996)]{klette1996inconsistency}
Tor~Jakob Klette and Zvi Griliches.
\newblock The inconsistency of common scale estimators when output prices are
  unobserved and endogenous.
\newblock \emph{Journal of Applied Econometrics}, pages 343--361, 1996.

\bibitem[Larochelle et~al.(2012)Larochelle, Mandel, Pascanu, and
  Bengio]{larochelle2012learning}
Hugo Larochelle, Michael Mandel, Razvan Pascanu, and Yoshua Bengio.
\newblock Learning algorithms for the classification restricted boltzmann
  machine.
\newblock \emph{Journal of Machine Learning Research}, 13\penalty0
  (Mar):\penalty0 643--669, 2012.

\bibitem[Morikawa et~al.(2002)Morikawa, Ben-Akiva, and
  McFadden]{morikawa2002discrete}
Taka Morikawa, Moshe Ben-Akiva, and Daniel McFadden.
\newblock Discrete choice models incorporating revealed preferences and
  psychometric data.
\newblock In \emph{Advances in Econometrics}, pages 29--55. Emerald Group
  Publishing Limited, 2002.

\bibitem[Paulssen et~al.(2014)Paulssen, Temme, Vij, and
  Walker]{paulssen2014values}
Marcel Paulssen, Dirk Temme, Akshay Vij, and Joan~L Walker.
\newblock Values, attitudes and travel behavior: a hierarchical latent variable
  mixed logit model of travel mode choice.
\newblock \emph{Transportation}, 41\penalty0 (4):\penalty0 873--888, 2014.

\bibitem[Rosenfeld and Kraus(2009)]{rosenfeld2009modeling}
Avi Rosenfeld and Sarit Kraus.
\newblock Modeling agents through bounded rationality theories.
\newblock In \emph{IJCAI}, volume~9, pages 264--271, 2009.

\bibitem[Rosenfeld et~al.(2012)Rosenfeld, Zuckerman, Azaria, and
  Kraus]{rosenfeld2012combining}
Avi Rosenfeld, Inon Zuckerman, Amos Azaria, and Sarit Kraus.
\newblock Combining psychological models with machine learning to better
  predict people's decisions.
\newblock \emph{Synthese}, 189\penalty0 (1):\penalty0 81--93, 2012.

\bibitem[Sobhani and Farooq(2017)]{sobhani2017innovative}
A.~Sobhani and B.~Farooq.
\newblock Innovative intercity transport mode: Application of choice preference
  integrated with attributes nonattendance and value learning.
\newblock In \emph{21st International Federation of Operational Research
  Societies, Québéc City}, 2017.

\bibitem[Vij and Walker(2016)]{vij2016and}
Akshay Vij and Joan~L Walker.
\newblock How, when and why integrated choice and latent variable models are
  latently useful.
\newblock \emph{Transportation Research Part B: Methodological}, 90:\penalty0
  192--217, 2016.

\bibitem[Wong et~al.(2017)Wong, Farooq, and Bilodeau]{wong2017discriminative}
Melvin Wong, Bilal Farooq, and Guillaume-Alexandre Bilodeau.
\newblock Discriminative conditional restricted boltzmann machine for discrete
  choice and latent variable modelling.
\newblock \emph{Journal of Choice Modelling}, 2017.

\end{thebibliography}

\end{document}